\documentclass[conference]{IEEEtran}

  	\usepackage[pdftex]{graphicx}
  	\graphicspath{{../pdf/}{../jpeg/}}
	\DeclareGraphicsExtensions{.pdf,.jpeg,.png}

	\usepackage[cmex10]{amsmath}
	\usepackage{mathabx}
	\usepackage{algorithmic}
	\usepackage{array}
	\usepackage{mdwmath}
	\usepackage{mdwtab}
	\usepackage{eqparbox}
	\usepackage{url}
\usepackage[utf8]{inputenc}
\usepackage[T2A]{fontenc}
\usepackage[ruled]{algorithm2e}
\begin{document}

\title{\LARGE Efficient Chemical Space Exploration Using Active Learning Based on Marginalized Graph Kernel: an Application for Predicting the Thermodynamic Properties of Alkanes with Molecular Simulation}


 \author{\authorblockN{Yan Xiang\authorrefmark{2}, Yu-Hang Tang\authorrefmark{3}, Zheng Gong\authorrefmark{2}, Hongyi Liu\authorrefmark{2}, Liang Wu\authorrefmark{2}, Guang Lin\authorrefmark{4},\authorrefmark{1} Huai Sun\authorrefmark{2},\authorrefmark{1}}
 \authorblockA{\authorrefmark{2}School of Chemistry and Chemical Engineering, Shanghai Jiao Tong University, Shanghai 200240, China}
 \authorblockA{\authorrefmark{3}Lawrence Berkeley National Laboratory, Berkeley, California 94720, United States}
 \authorblockA{\authorrefmark{4}Department of Mathematics \& School of Mechanical Engineering, \\Purdue University, West Lafayette, Indiana 47907, United States}
  \authorblockA{\authorrefmark{1}Corresponding to huaisun@sjtu.edu.cn, guanglin@purdue.edu}}

\maketitle
\thispagestyle{plain}
\pagestyle{plain} 
\begin{abstract}
We introduce an explorative active learning (AL) algorithm based on Gaussian process regression and marginalized graph kernel (GPR-MGK) to explore chemical space with minimum cost. Using high-throughput molecular dynamics simulation to generate data and graph neural network (GNN) to predict, we constructed an active learning molecular simulation framework for thermodynamic property prediction. In specific, targeting 251,728 alkane molecules consisting of 4 to 19 carbon atoms and their liquid physical properties: densities, heat capacities, and vaporization enthalpies, we use the AL algorithm to select the most informative molecules to represent the chemical space. Validation of computational and experimental test sets shows that only 313 (0.124\% of the total) molecules were sufficient to train an accurate GNN model with $\rm R^2 > 0.99$ for computational test sets and $\rm R^2 > 0.94$ for experimental test sets. We highlight two advantages of the presented AL algorithm: compatibility with high-throughput data generation and reliable uncertainty quantification.
\end{abstract}

\IEEEoverridecommandlockouts
\begin{keywords}
Active Learning, Marginalized Graph Kernel, Thermodynamic Property
\end{keywords}

\IEEEpeerreviewmaketitle


\section{Introduction}
Thermodynamic properties of molecular liquids play an important role in chemical engineering. They can be obtained through either experimental or computational means. For example, NIST ThermoData Engine (TDE) collected experimental data from literature and provides critically evaluated thermodynamic data based on a large variety of available models \cite{diky_thermodata_2016}. Molecular simulation (MS) offers another opportunity to calculate the thermodynamic properties of molecular liquids from scratch. The Industrial Fluid Properties Simulation Challenges (IFPSC) \cite{noauthor_industrial_nodate, case_first_2004} demonstrated that computational approaches have the potential to predict thermodynamic and transport properties at the level of experimental reproducibility. However, both experimental and computational methods require high costs, and only limited data can be generated in practice. Machine learning (ML) is believed to have great potential to revolutionize the process of chemical discovery, thus expanding the exploration of chemical compound space (CCS) by orders of magnitude \cite{vonlilienfeld_quantum_2018, von_lilienfeld_exploring_2020, tkatchenko_machine_2020, duan_putting_2021, zhong_accelerated_2020, lu_neural_2020, chen_machine-learning-accelerated_2021, nandy_strategies_2018}.

Active learning (AL) are iterative algorithms that attempt to maximize the ML model’s performance with minimal data generation \cite{settles_active_2009, mackay_information-based_1992, cohn_active_1996}. In the context of this article, we focus on the AL of CCS, i.e., how to find a representative subset from a large number of candidate molecules. At each loop of AL, a surrogate ML model is trained using the known data, and the property-unknown candidate molecules are predicted. According to the predictions and acquisition function, one or more of the most valuable molecules are selected and their properties are generated through experimental or computational approaches. When the acquisition function is \textit{predicted uncertainty}, the selected samples will cover the entire CCS and the surrogate model will achieve the highest predictive performance, known as \textit{uncertainty reduction} or \textit{explorative active learning}. When the acquisition function is the \textit{predicted value}, AL prefers to select molecules with desired properties rather than exploring unknown regions, which is called \textit{greedy} or \textit{exploitive active learning} \cite{reker_active-learning_2015, reker_practical_2019}.

Due to the rapid development of deep learning and graph neural networks (GNNs) \cite{zhou_graph_2020, gilmer_neural_2017, kipf_semi-supervised_2017, velickovic_graph_2018} recently, research on how to use AL to explore CCS has gained renewed attention, especially in drug discovery. For example, Zhang and Lee proposed Bayesian semi-supervised graph convolutional neural networks to achieve uncertainty quantification (UQ) and AL \cite{zhang_bayesian_2019}; Graff et al. explored the application of batch active learning for high-throughput virtual screening \cite{graff_accelerating_2021}. They used directed message-passing neural network (D-MPNN) \cite{yang_analyzing_2019} as the surrogate model and mean-variance estimation (MVE) \cite{hirschfeld_uncertainty_2020} as the UQ method; Soleimany et al. demonstrated that evidential deep learning outperforms MVE on UQ and AL tasks \cite{soleimany_evidential_2021}. However, one of the key challenges to address is the incompatibility between the sequential nature of AL algorithms and the parallel nature of high-throughput simulations or experiments. In fact, most experimental instruments and equipment are designed to test multiple samples simultaneously, not one by one. Batch AL solves this problem by selecting multiple samples in each AL iteration, but at the cost that the selected samples contain redundant information and the performance is significantly worse than one-by-one AL \cite{zhang_bayesian_2019, graff_accelerating_2021}.

In this work, we developed an explorative AL to find a representative subset of the target CCS to maximize prediction accuracy and minimize the computational cost. We use Gaussian process regression coupled with marginalized graph kernel, abbreviated as GPR-MGK \cite{tang_prediction_2019, xiang_predicting_2021, xiang_comparative_2021}, as the surrogate ML model, and take advantage of the method in uncertainty quantification and compatibility with parallel high-throughput simulations. Together with the high-throughput molecular simulation (HT-MS) engine for automatically generating data (based on the TEAM-AMBER force field and simulation protocols developed by Gong et al. \cite{gong_temperature_2018, gong_predicting_2018}) and D-MPNN, it is possible to predict the properties of the entire CCS. The whole procedure is automated and is called active learning molecular simulation (ALMS) workflow.

In specific, we targeted the CCS of all alkane molecules with the number of carbon atoms ranging from four to nineteen ($4\leq N\leq 19$), which contains 251,728 molecules. By setting the uncertainty thresholds of 0.5, 0.4, and 0.3, we selected 313, 599, and 1679 molecules as the training sets respectively. We validated our model by evaluating the prediction error on both computational and experimental test sets. The computational test set contains 3,000 molecules randomly selected from the CCS excluding those selected by AL, and the experimental test set contains 240 molecules collected from the NIST database \cite{diky_thermodata_2016, noauthor_knovel_nodate}. Although still at an early stage, our study indicates the possibility of obtaining large-scale thermodynamic properties through active machine learning and molecular dynamics simulations. And it is worth noting that the computational cost does not increase exponentially with molecular size as the number of molecules in CCS. We believe that MGK has a promising future in molecular active learning due to its compatibility with high-throughput simulations or experiments and accurate uncertainty quantification.


\section{METHODS}
\begin{figure}[ht!] 
\centering
\includegraphics[width=3.4in]{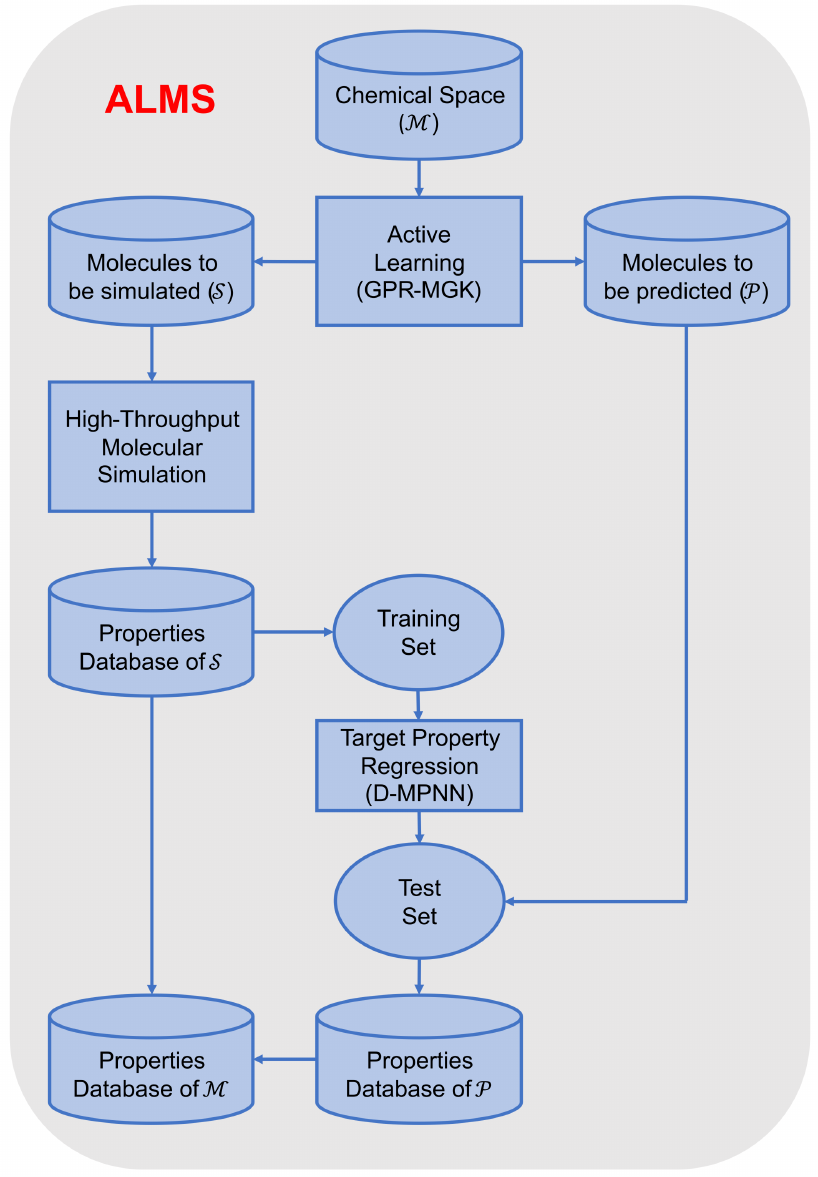}
\caption{Overview of the active learning molecule simulation workflow. The framework starts with a set of molecules $\mathcal{M}$ as input, whose properties are the targets. An active learning algorithm is used to select a representative subset $\mathcal{S} \in \mathcal{M}$ for molecular simulation. The selected molecules $\mathcal{S}$ are then fed into the high-throughput molecular simulation module for data production. A D-MPNN is trained based on the simulation data and makes predictions on the unselected molecules $\mathcal{P}$. Finally, a database containing the thermodynamic properties of $\mathcal{M}$ is obtained.}
\label{f1}
\end{figure}

Fig. \ref{f1} illustrates the ALMS workflow. The starting point is a set of candidate molecules $\mathcal{M}$. Through explorative AL, $\mathcal{M}$ is divided into two subsets $\mathcal{S}$ and $\mathcal{P}$. The subset $\mathcal{S}$ contains the molecules to be simulated and the subset $\mathcal{P}$ contains the molecules to be predicted. Then the subset $\mathcal{S}$ is sent to an HT-MS engine to generate a molecular property database through quantum chemistry calculation and molecular dynamics simulation. Finally, these MS data are used to train an ML regressor, which can accurately predict the properties of the subset $\mathcal{P}$, and an ML database containing the properties of $\mathcal{M}$ is obtained.

\subsection{Explorative Active Learning}
The AL algorithm used in this work is shown in Algorithm \ref{a1}. At the beginning, the selected set $\mathcal{S}$ (two molecules) are randomly selected from the target CCS $\mathcal{M}$, where the remaining molecules constitute the pool set $\mathcal{P}$ to be selected during the AL process. In each iteration of AL, a Gaussian process regression (GPR) \cite{rasmussen_gaussian_2006} model is trained using the selected set $\mathcal{S}$ and used to predict the uncertainties of $\mathcal{P}$

\begin{equation}
\label{eq:gpr}
\rm {\bf U}_P = diag({\bf K}_{\rm PP} - {\bf K}_{\rm SP}^{T}{\bf K}_{\rm SS}^{-1} {\bf K}_{\rm SP})
\end{equation}
where ${\bf K}_{\rm PP}$ is the kernel matrix of $\mathcal{P}$, ${\bf K}_{\rm SS}$ is the kernel matrix of $\mathcal{S}$, ${\bf K}_{\rm SP}$ is the kernel matrix between $\mathcal{S}$ and $\mathcal{P}$. The molecule $\alpha$ with the highest predicted uncertainty is then added to the selected set $\mathcal{S}$. In practice, the pool set $\mathcal{P}$ may contain a large number of molecules, so it is too time-consuming to calculate the predicted uncertainty of all molecules in the pool set. We accelerated the active learning process by randomly selecting $N_t$ molecules from $\mathcal{P}$ for prediction, discarding all data points with prediction uncertainty less than the threshold $U_t$, and then selecting the one with the highest prediction uncertainty to add to $\mathcal{S}$. AL iterations are repeated until the pool set $\mathcal{P}$ is empty (all predicted uncertainties of P are less than the threshold value $U_t$). The AL algorithm described above is called explorative AL because the criterion for selecting the most valuable molecules is the predicted uncertainty, i.e., the AL algorithm keeps exploring the CCS by finding the molecule in $\mathcal{P}$ that is least similar to $\mathcal{S}$. Therefore, $\mathcal{S}$ is a representative subset of the CCS $\mathcal{M}$ at the end of AL. Every molecule in $\mathcal{S}$ always has similar molecules in S, which guarantees that an ML model trained using $\mathcal{S}$ as a training set can make accurate predictions for all molecules in $\mathcal{P}$. Furthermore, the molecules in $\mathcal{S}$ are not similar to each other, which indicates the number of molecules in $\mathcal{S}$, as well as the cost of obtaining the target properties of these molecules, are minimized.

\begin{algorithm}
    \SetAlgoLined
    \KwIn{\\$\mathcal{M}$: target CCS; \\ $N_t$: number of samples to be predicted in each active learning iteration;\\ $U_t$: uncertainty threshold.}
    \KwOut{$\mathcal{S}$: Selected set }
    Randomly select 2 samples from $\mathcal{M}$ as initial $\mathcal{S}$, the rest samples are pool set $\mathcal{P}$\;
    $\mathcal{A} \gets \emptyset$: Abandoned set\;
    \While{$\mathcal{P} \neq \emptyset$}{
        \eIf{$|\mathcal{P}| > N_t$}{
            Randomly select a subset $\mathcal{T} \in \mathcal{P}, |\mathcal{T}|=N_t$\;
            }{
            $\mathcal{T} \gets \mathcal{P}$\;
        }
        $\rm {\bf U}_T \gets diag({\bf K}_{TT}-{\bf K}^{T}_{ST} {\bf K}^{-1}_{TT} {\bf K}_{ST})$: Predicted uncertainty\;
        \If{$\rm max({\bf U}_T) > {\mit U_t}$}{
            $\alpha \gets \rm argmax({\bf U}_T)$\;
            $\mathcal{S} \gets \mathcal{S} \cup \left\{\alpha\right\}$\;
            $\mathcal{P} \gets \mathcal{P} \space \backslash \left\{\alpha\right\}$\;
        }
        $\{\beta_i, i=1,2,\cdots\} \gets \rm arg({\bf U}_T<{\mit U_t})$\;
        $\mathcal{A} \gets \mathcal{A} \cup \{\beta_i, i=1,2,\cdots\}$\;
        $\mathcal{P} \gets \mathcal{P} \space \backslash \{\beta_i, i=1,2,\cdots\}$\;
    }
\caption{Explorative Active Learning using GPR-MGK}
\label{a1}
\end{algorithm}

We chose GPR-MGK as the surrogate ML model to compute the predicted uncertainty as shown in equation \ref{eq:gpr}. The MGK was used to compute the kernel matrix describing the similarity between molecules by comparing the similarity between simultaneous random walk paths on a pair of graphs. We used the MGK proposed in our previous work \cite{xiang_predicting_2021}, which is designed to predict the thermodynamic properties of molecular liquids. The formula of MGK is expressed as:

\begin{align}
K(G,& G^\prime) = \sum_{l=1}^{\infty} \sum_{\bf h} \sum_{\bf h^{\prime}} \left[ p_s\left(h_1\right) p^{\prime}_s\left(h^{\prime}_1\right) K_v\left(v_{h_1}, v^{\prime}_{h_1^{\prime}}\right) p_q\left(h_l\right)\right.\nonumber \\ 
\times &\left. p_q^{\prime}\left(h_l^{\prime}\right) \left(\prod_{i=2}^l p_t(h_i|h_{i-1}) \right) \left(\prod_{j=2}^l p^{\prime}_t(h^{\prime}_i|h^{\prime}_{i-1}) \right)\right.\nonumber \\ 
\times &\left. \left( \prod_{k=2}^l K_v\left(v_{h_k},v^{\prime}_{h^{\prime}_k}\right)K_e\left(e_{h_k,h_{k-1}},e^{\prime}_{h^{\prime}_k,h^{\prime}_{k-1}}\right) \right) \right],
\label{eq:mgk}
\end{align}

\noindent where $G$ and $G^{\prime}$ are the two graphs (molecules) to be computed; $\bf h$ and $\bf h^\prime$ are the random walk paths of length $l$; $p_s,p_q,p_t$ are the starting probability, stopping probability, and transition probability, of the random walk process, respectively; The atom kernel $K_v(\cdot, \cdot)$computes the similarity between a pair of atoms as the product of atomic feature kernels:
\begin{equation}
    K_v\left(v,v^{\prime}\right) = \prod_j\delta_j\left(\phi_j\left(v\right),\phi_j\left(v^{\prime}\right)\right),
\end{equation}
where the atomic feature kernel $\delta_j(\cdot, \cdot)$ computes the similarity between a pair of atomic features, usually using a Kronecker delta function
\begin{equation}
\delta\left(\phi_1, \phi_2\right) = \left\{
  \begin{array}{ c l }
    1 & \quad \textrm{if } \phi_1 = \phi_2 \\
    h \in (0,1) & \quad \textrm{otherwise}.
  \end{array}
\right.
\end{equation}

The bond kernel $K_e(\cdot, \cdot)$ is defined in the same way as atom kernel. In addition, the MGK is normalized to improve prediction performance
\begin{align}
\bar{K}\left(G,G^{\prime}\right) = & \frac{K\left(G,G^{\prime}\right)}{\sqrt{K\left(G,G\right)K\left(G^{\prime},G^{\prime}\right)}} \times \nonumber \\
&{\rm exp} \left[ -\frac{\left(K\left(G,G\right)-K\left(G^{\prime},G^{\prime}\right)\right)^2}{\lambda^2}\right].
\end{align}
The atom features, bond features, and hyperparameters were tuned by minimizing the cross-validation error in the critical temperature data set. For the details about MGK, we refer the reader to ref \cite{xiang_predicting_2021}.

Here, we highlight the advantages of GPR-MGK in AL. The first advantage of GPR-MGK is its reliable uncertainty quantification, as demonstrated in our previous studies \cite{xiang_predicting_2021, xiang_comparative_2021}. The quality of the surrogate model’s predicted uncertainty is critical to the AL algorithm because it guarantees that the selected sample is truly different from the molecules in $\mathcal{S}$, resulting in maximum information gain. The second advantage is the compatibility with high-throughput simulations or experiments, i.e., data selection can be performed prior to data generation. According to our previous studies, the hyperparameters optimized using the critical temperature data set work well in various data sets such as density, heat capacity, viscosity, etc\cite{xiang_predicting_2021}. The transferability of hyperparameters allows us to use the same hyperparameters during the AL process. According to equation \ref{eq:gpr}, the predicted uncertainties of GPR only depend on the kernel matrices, which are computed from the molecular graphs, and their target properties of the molecules in S are not required. Therefore, the selected subset $\mathcal{S}$ can be obtained prior to data generation using MS. The logic behind this is transfer learning \cite{cai_transfer_2020}, where the hyperparameters are the “knowledge” learned from critical temperature data sets that can be transferred to other data sets. In contrast, with graph neural networks or random forests as surrogate models, each AL iteration must wait for the MS of the newly added molecules in the previous iteration to complete before proceeding. Batched active learning that add multiple samples in each AL iteration is a common solution for this compatibility \cite{zhang_bayesian_2019, graff_accelerating_2021}, but the performance suffers severely due to the high similarity between the samples added in each batch.

\subsection{High-Throughput Molecular Simulation}
We used the TEAM-AMBER force field \cite{gong_temperature_2018} and HT-MS developed by Gong et al. \cite{gong_predicting_2018, cao_high-throughput_2017} to generate the liquid density, heat capacity, and heat of vaporization (HOV) of alkane liquids. Liquid density was computed by averaging the density of the trajectories. Heat capacity was computed by combining the intramolecular and intermolecular contributions
\begin{equation}
    C_P = \overbrace{C_{\rm trans} + C_{\rm rot} + C_{\rm vib}}^{\rm intra} + \overbrace{\left(\frac{dU_{\rm inter}}{dT}\right)_P + \left(\frac{PdV}{dT}\right)_P}^{\rm inter},
\end{equation}
where $C_{\rm trans}$ and $C_{\rm rot}$ are the ideal translational and rotational contributions computed through the equipartition theorem, $C_{\rm vib}$ is the vibrational contribution computed using the hindered-rotor model \cite{ayala_identification_1998}, and $U_{\rm inter}$ is the intermolecular potential. In literature \cite{sun_compass_1998, wang_application_2011}, HOV was broadly computed as
\begin{equation}
\label{eq:hov}
    H_{\rm vap} = RT - U_{\rm inter}.
\end{equation}

Gong et al. found that equation \ref{eq:hov} systematically overestimates the HOV of alkanes, thus proposing an empirical correction
\begin{equation}
    C_{\rm e} = - \frac{1}{15} n_{C}RT,
\end{equation}
where $n_C$ is the number of carbon atoms, and the coefficient $-\frac{1}{15}$ was obtained by fitting the experimental data.

For each molecule, the simulated temperature was set to 16 data points ranging from 0.4 $T_c$ to 0.9 $T_c$, and the simulated pressure was set to 1 bar. The critical temperature $T_c$ was predicted by a D-MPNN \cite{yang_analyzing_2019} trained on experimental data from the NIST database \cite{diky_thermodata_2016, noauthor_knovel_nodate}. With the SMILES string, temperature, and pressure as inputs, the HS-MS module automatically generated all input files, submitted the jobs to the HPC clusters, analyzed the simulation trajectory, and collected the results. OpenBabel \cite{oboyle_open_2011} and Packmol \cite{martinez_packmol_2009} were used to generate input 3D molecular configurations. The quantum chemistry calculations were performed using Gaussian 16 \cite{g16}. The molecular dynamics simulations were performed using GROMACS \cite{hess_gromacs_2008}. We developed a series of quality control tests, and simulations that fail were discarded or extended until they pass the test: (1) Check the distribution of the kinetic energy trajectory via the Kolmogorov-Smirnov test. The null hypothesis is the distribution satisfies the Maxwell-Boltzmann distribution, and the simulation is extended if p-value less than 0.01 \cite{merz_testing_2018}. (2) If the density is less than 50 $\rm kg \cdot m^{-3}$, the simulation is considered as vapor and discarded. (3) If the diffusion coefficient is less than $\rm 10^{-8} cm^2 \cdot s^{-1}$, the simulated system is considered solid and discarded. (4) The trajectories of temperature, pressure, density, and potential are detected using the method proposed by Chodera to determine the convergence and production interval \cite{chodera_simple_2016}. (5) If the density, heat capacity, and HOV are not monotonic functions of temperature, the simulations are discarded. (6) The density, heat capacity, and HOV are fitted as quadratic functions of temperature. If the fitting score $\rm R^2$ is less than 0.98, the simulations are discarded. For more details on the HT-MS, we refer the reader to ref \cite{gong_predicting_2018}.

\subsection{Property Prediction using D-MPNN}
After the MS was completed, an MS database $\left\{ \left( x_i,T_i,\rho_i,C_{p,i},H_{\rm vap, \mit i}\right), i=1,2, \cdots \right\}$ was generated, where $x_i \in \mathcal{S}$ is a molecule selected via AL, $T_i$ is the temperature, $\rho$ is the density, $C_{p,i}$ is heat capacity and $H_{\rm vap, \mit i}$ is HOV. Pressure is not considered because the pressure was set to 1 bar for all simulations. Three D-MPNNs were trained using $\left\{x,T\right\}$ as input and $\rho, C_P, H_{\rm vap}$ as output respectively. The density, heat capacity, and HOV of all molecules in $\mathcal{M}$ were predicted using the D-MPNNs.

D-MPNN \cite{yang_analyzing_2019} is a variant of message passing neural network (MPNN) \cite{gilmer_neural_2017} which predicts molecular properties with molecular graphs and RDKit features \cite{noauthor_descriptor_2021} as input. D-MPNN contains two phases: message passing and readout. In the message-passing phase, messages are passed through bonds, and the hidden states of atoms and bonds are updated according to the incoming messages. The hidden states of atoms are aggregated (sum or average) to obtain a feature vector of the molecular graph. The molecular representation is the concatenation between the feature vector learned via message passing and the RDKit feature vector \cite{noauthor_descriptor_2021}. The readout phase is a fully connected neural network that connects the learned molecular representation from message passing and target property. We used Chemprop \cite{noauthor_message_2021} library for training and prediction of D-MPNN with default settings except for followings: 100 epochs for training; aggregation function was “mean” for density and HOV, “sum” for heat capacity. In order to incorporate temperature into the model input, we concatenated molecular representation with temperature between the message-passing and readout phases.

We chose D-MPNN because its computational cost is lower than GPR-MGK. Our previous benchmark study shows that the predictions of D-MPNN are correlated to that of the GPR-MGK at a per-sample level \cite{xiang_comparative_2021}. Therefore, it is reliable to first use GP-MGK AL to select samples to construct the training set, and then use D-MPNN to make predictions.

\subsection{Molecule Enumeration}
We used alkane with 4 to 19 heavy atoms as target CCS. The SMILES strings for these molecules were generated by an atom-addition algorithm. Alkanes with N heavy atoms were generated by adding a carbon atom to the alkanes with N-1 heavy atoms. Duplicate SMILES strings were discarded by canonicalization using RDKit\cite{landrum_rdkit_2013}. In total, we collected 251728 distinct molecules.


\section{Results and Discussion}

\subsection{Explorative Active learning}
The target CCS contains 251728 molecules. We performed a three-stage AL with uncertainty thresholds $U_t=0.5,0.4,0.3$, and three subsets $\mathcal{S}_{U_t=0.5}$ (313 molecules), $\mathcal{S}_{U_t=0.4}$ (599 molecules), $\mathcal{S}_{U_t=0.3}$ (1679 molecules) were obtained, respectively. The AL set with a high uncertainty threshold is the subset of the AL set with a low uncertainty threshold: $\mathcal{S}_{U_t=0.5} \subset \mathcal{S}_{U_t=0.4} \subset \mathcal{S}_{U_t=0.3}$. 

\begin{figure}[ht!] 
\centering
\includegraphics[width=3.4in]{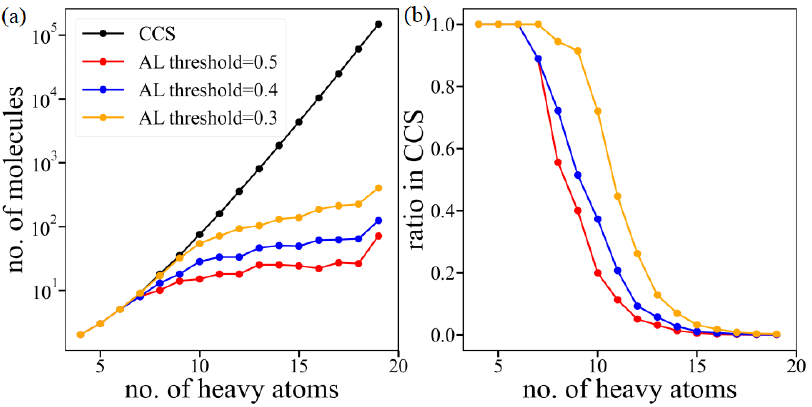}
\caption{(a) The relationship between the number of molecules and the number of heavy atoms for the CCS and AL sets. (b) The relationship between the ratio of the number of molecules in the AL set to the CCS and the number of heavy atoms.}
\label{fig2}
\end{figure}

The distribution of the CCS and AL sets along the number of heavy atoms is shown in Figure \ref{fig2}. The number of molecules in the CCS grows exponentially with heavy atoms, while the number of molecules selected by AL grows much more slowly. This suggests that the distribution of molecules in CCS becomes denser as the number of heavy atoms increases, so the proportion of molecules selected by AL becomes less and less. Therefore, it is promising to explore a huge CCS with affordable cost by only focusing on the small subset selected by AL. As elucidated below, the ML model trained with only hundreds of molecules selected by AL predicts the entire CCS with extremely high accuracy.

\subsection{Prediction Performance-Simulation Test Set}
The densities, heat capacities, and HOV of the AL-selected molecules were obtained through HT-MS, and D-MPNNs were trained for each property and each AL set respectively. In order to evaluate the prediction accuracy of the ALMS framework, a test set $\mathcal{S}_{\rm test}$ (3000 molecules) were randomly selected from the CCS excluding the AL sets.

For different AL sets as training sets and different target properties, the predictions of D-MPNNs versus the simulation data of the test set are compared in Figure 3. For $\mathcal{S}_{U_t=0.5}$, the root mean square errors (RMSEs) of the predictions are 6.2 $\rm kg \cdot m^{-3}$ for density, 9.7 $\rm J \cdot mol^{-1} \cdot K^{-1}$ for heat capacity and 0.94 $\rm kJ \cdot mol^{-1}$ for HOV. Despite the high accuracy, $\rm R^2>0.99$, continued improvement in prediction accuracy was observed as $U_t$ decreased to 0.4, 0.3, and the training set increased from 313 to 599, 1679 molecules.
\begin{figure}[ht!] 
\centering
\includegraphics[width=3.4in]{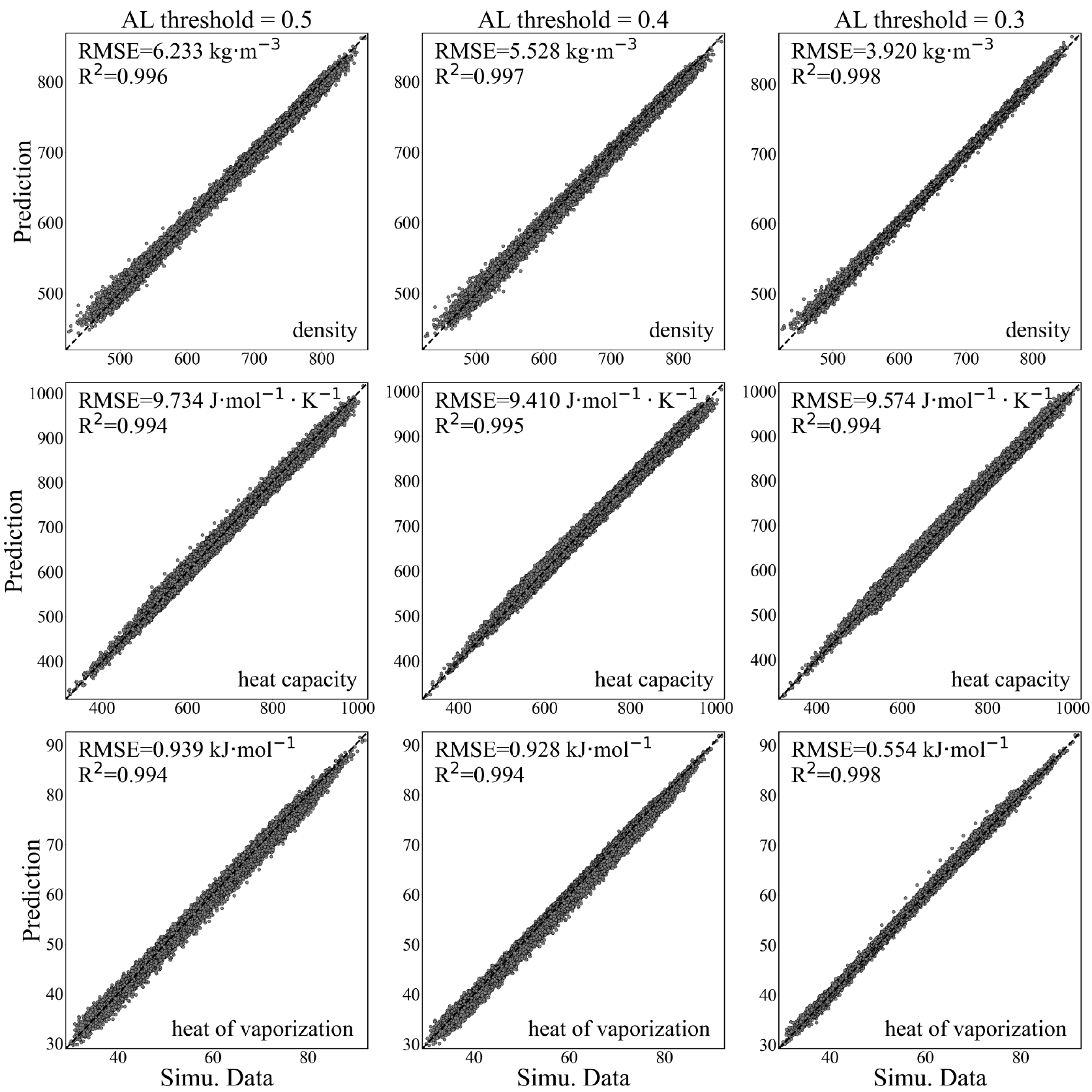}
\caption{Comparison of D-MPNN predictions with simulation data. The columns show that the training sets are selected through active learning with thresholds 0.5, 0.4, 0.3. The rows show the target properties: density, heat capacity, and heat of vaporization.}
\label{fig3}
\end{figure}

The uncertainty threshold $U_t$ in AL is a parameter that balances the computational cost and prediction accuracy. The smaller the $U_t$, the more molecules are simulated, and the larger the computational cost, the more accurate the ALMS prediction. For the CCS of alkane considered in this work, setting $U_t$ to 0.5 is accurate enough for practical consideration. Further reduction of $U_t$ requires several times the computational cost, but the improvement in model prediction accuracy is relatively small.

\subsection{Active Learning VS Random Sampling}
\begin{figure}[ht!] 
\centering
\includegraphics[width=3.4in]{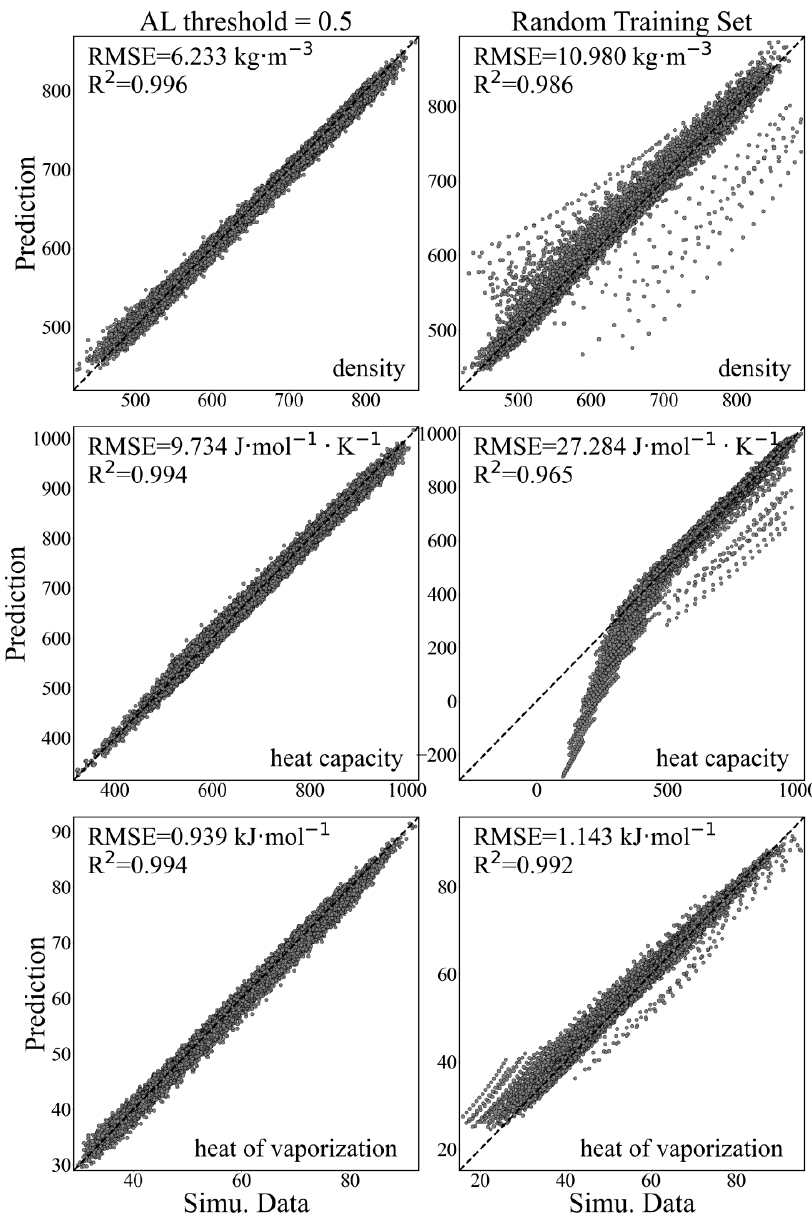}
\caption{Comparison of D-MPNN predictions with simulation data. Left: the training sets are selected through active learning with uncertainty threshold of 0.5. Right: the training sets are randomly selected. The rows show the target properties: density, heat capacity, and heat of vaporization.}
\label{fig4}
\end{figure}

To evaluate the power of the explorative AL algorithm introduced in this study, we conducted a controlled experiment. 313 molecules (the same as $\mathcal{S}_{U_t=0.5}$) are randomly selected from $\mathcal{S}_test$ as the training set of D-MPNN, and the union of the remaining molecules and $\mathcal{S}_{U_t=0.5}$ is used as the test set. Although this is not a truly random sample, we believe it has no effect on our conclusions, as only $\frac{313}{251728} = 0.667\%$ of the molecules is ignored. For all target properties, AL outperformed random sampling significantly as indicated in Figure \ref{fig4}. The drawback of random sampling is that it lacks the sampling of sparse regions in the CCS. In our study case, an aspect easily understood is that alkanes with heavy atomic numbers from four to six were all selected by AL, because these molecules are small and they sparsely distributed on the CCS.

\subsection{Prediction Performance-Experiment Test Set}
\begin{figure}[ht!] 
\centering
\includegraphics[width=3.4in]{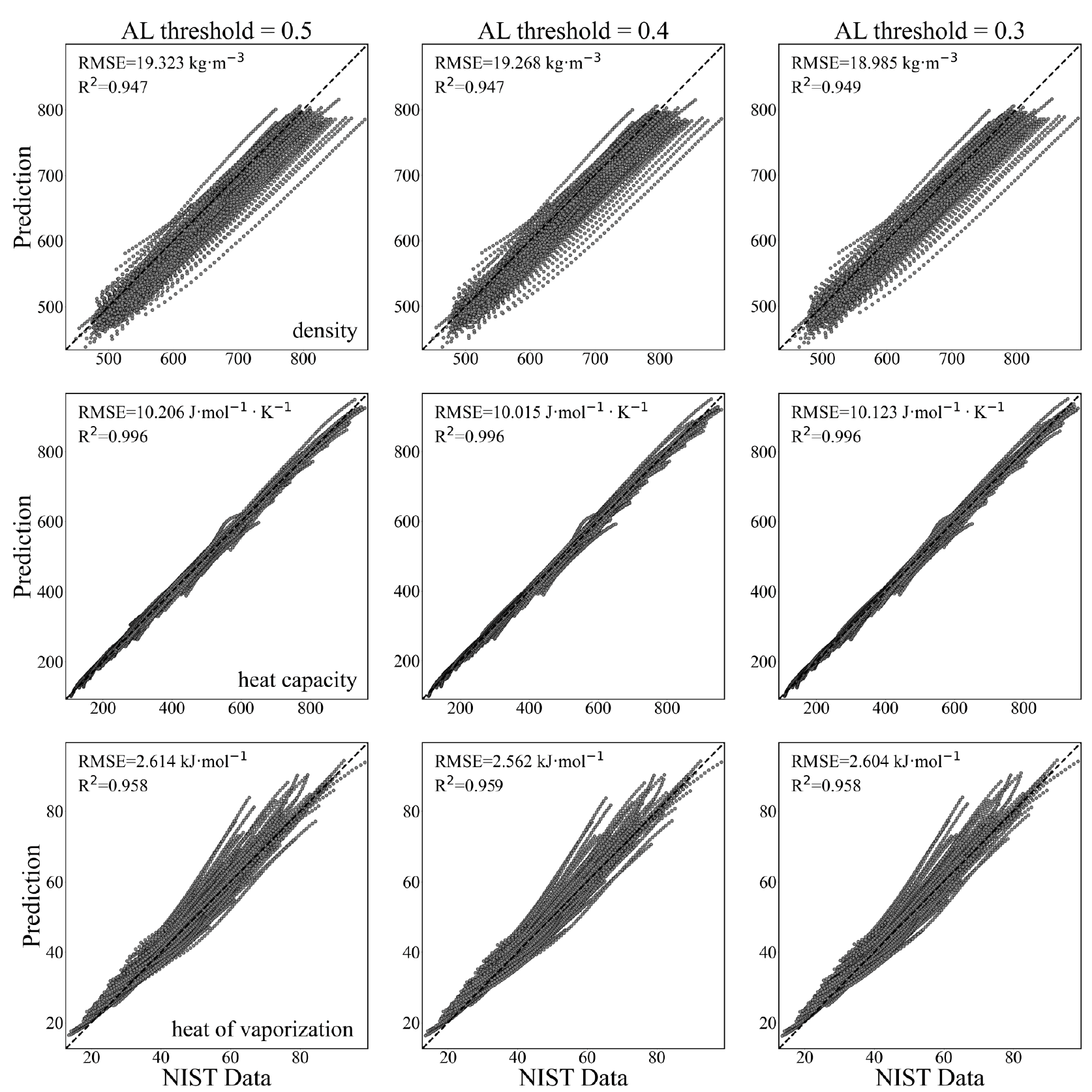}
\caption{Comparison of D-MPNN predictions with NIST data. The three columns indicate different training sets, obtained by active learning with thresholds of 0.5, 0.4 and 0.3, respectively. The three rows represent the three target properties: density, heat capacity and heat of vaporization.}
\label{fig5}
\end{figure}

It is essential to evaluate our model by comparing the predictions with experimental data, which examines the combinatory prediction errors due to both machine learning and MS. From the NIST database collected by knovel \cite{diky_thermodata_2016, noauthor_knovel_nodate}, experimental data on the density, heat capacity, and HOV of 240, 240, and 235 alkanes were collected, respectively. For different AL sets as training sets and different target properties, the predictions of D-MPNNs versus the experimental data of the test set are compared in Figure \ref{fig5}. For $\mathcal{S}_{U_t=0.5}$, The RMSEs and R2 are 19.3 $\rm kg\cdot m^{-3}$ and 0.947 for density, 10.2 $\rm J\cdot mol^{-1}\cdot K^{-1}$ and 0.996 for heat capacity and 2.6 $\rm kJ\cdot mol^{-1}$ and 0.958 for HOV. The RMSEs for density and HOV are about twice as high as when the test set was simulated data (Figure \ref{fig3}), indicating that the density and HOV calculated by MS, although accurate, still did not reach the level of experimental reproducibility. On the contrary, the predictions on heat capacity reach the level of experimental reproducibility. Another important phenomenon is that the prediction accuracy of $\mathcal{S}_{U_t=0.5}$, $\mathcal{S}_{U_t=0.4}$, $\mathcal{S}_{U_t=0.3}$ is almost the same, which indicates that simulating more molecules does not improve the prediction accuracy. This is because, simulating more molecules can only reduce the error of the ML but the error of MS dominates. Therefore, only a very small number of molecules need to be simulated to reach the upper limit of the ALMS predictions.

\section{Conclusion}
In this work, we developed an ALMS framework to predict the properties of a large number of molecules with a small computational cost. It consists of an explorative AL module for molecules selection, an HT-MS module for data generation, and an ML regression module for property prediction. Taking alkanes (4-19 heavy atoms, 251,728 molecules) as the target chemical space, the density, heat capacity and HOV of all molecules in the chemical space were accurately predicted by simulating only 313 (0.124\%) molecules.

The novelty of this work lies in explorative AL algorithm using GPR-MGK as the surrogate model for uncertainty quantification, which exhibits two advantages: (1) It is compatible with high-throughput simulations (or experiments), and all selected molecules are guaranteed to be dissimilar to each other, avoiding the redundant cost of duplication of information. (2) GPR-MGK model provides reliable uncertainty quantification. We believe that these advantages will bring a bright future for GPR-MGK in chemical space active learning.

Despite the success of this work, there are still many important issues that need to be addressed in the future. For example, in this work, the structure of MGK and the associated hyperparameters are fixed. How the hyperparameters of MGK affect the performance of AL has not been well investigated. In addition, benchmarking different AL algorithms is necessary for a more in-depth understanding.


\section*{Acknowledgment}
The computations in this paper were run on the $\pi$ 2.0 cluster supported by the Center for High-Performance Computing at Shanghai Jiao Tong University. This work was funded by the National Natural Science Foundation of China [Grant No. 21473112], [Grant No. 21403138], [Grant No. 21673138].

\section*{Data and Software Availability}
The code for alkane molecules generation is available at https://github.com/Xiangyan93/molecules-enumerate. The code for ALMS framework, and the simulation data are available at https://github.com/Xiangyan93/ALMS. The experimental data are taken from NIST database via Knovel at https://app.knovel.com/web/poc/ms/discovery.html.


\bibliographystyle{IEEEtran}
\bibliography{IEEEabrv,biblio_traps_dynamics}

\end{document}